\documentclass[sigconf,9pt,nonacm,balance]{acmart}

\AtBeginDocument{%
  \providecommand\BibTeX{{%
    \normalfont B\kern-0.5em{\scshape i\kern-0.25em b}\kern-0.8em\TeX}}}

\usepackage{microtype}
\usepackage{graphicx}
\usepackage{subfigure}
\usepackage{booktabs} 

\usepackage{hyperref}

\usepackage{enumitem}
\usepackage{amsmath}

\usepackage{balance}

\usepackage{xcolor}
\colorlet{BLACK}{black}
\colorlet{RED}{red}
\colorlet{BLUE}{blue}
\colorlet{PURPLE}{purple}
\newcommand\RevisionColor{black}
\newcommand\ReRevColor{black}
\newcommand\arXivColor{black}

\begin{document}

\title{A binary-activation, multi-level weight RNN and training algorithm for \textcolor{\RevisionColor}{ADC-/DAC-free and noise-resilient} processing-in-memory inference with eNVM} 


\author{Siming Ma}
\affiliation{\institution{Harvard University}}
\email{simingma@g.harvard.edu}

\author{David Brooks}
\affiliation{\institution{Harvard University}}
\email{dbrooks@eecs.harvard.edu}

\author{Gu-Yeon Wei}
\affiliation{\institution{Harvard University}}
\email{gywei@g.harvard.edu}

\newcommand{\setmuskip}[2]{#1=#2\relax}
\begingroup
\setmuskip{\thickmuskip}{-1mu}
\setmuskip{\medmuskip}{0mu}
\setmuskip{\thinmuskip}{-1mu}

\begin{abstract}
We propose a new algorithm for training neural networks with binary activations and multi-level weights, which enables efficient processing-in-memory circuits with embedded nonvolatile memories (eNVM). Binary activations obviate costly DACs and ADCs. Multi-level weights leverage multi-level eNVM cells. 
Compared to existing algorithms, our method not only works for feed-forward networks (e.g., fully-connected and convolutional), but also achieves higher accuracy and noise resilience for recurrent networks. 
In particular, we present an RNN-based trigger-word detection PIM accelerator, 
\textcolor{\RevisionColor}{
with detailed hardware noise models and circuit co-design techniques, and validate our algorithm's high inference accuracy and robustness against a variety of real hardware non-idealities.}
\end{abstract}




\maketitle

\begin{figure*}[h]
  \centering
	\includegraphics[width=\linewidth]{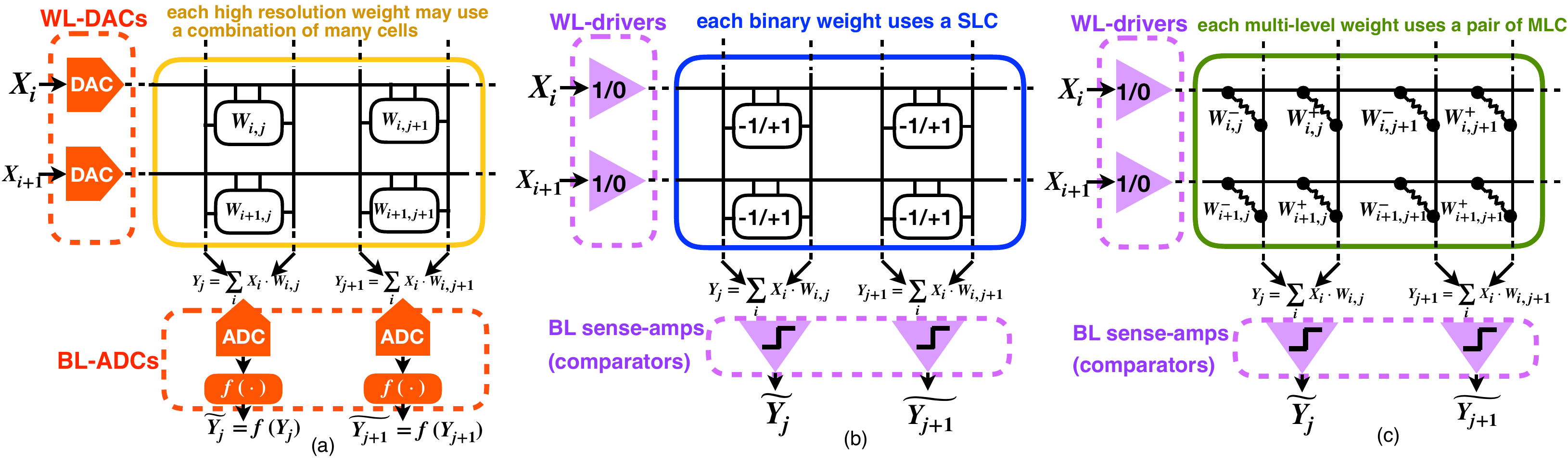} 
	\vspace{-6mm}
	\caption{(a) The conventional PIM architecture with DACs and ADCs for high-precision activations and a combination of many cells for each high-resolution weight. (b) an ADC-/DAC-free PIM architecture implementing BNNs, using SLCs for binary weights, and 
	(c) the optimal ADC-/DAC-free PIM architecture implementing our BA-MLW NNs, using a pair of MLCs for each MLW.} 
	\vspace{-3mm}
  \label{PIM}
\end{figure*}

\section{Introduction}
\label{intro}

Processing-in-memory (PIM) architectures for hardware neural network (NN) inference have gained increasing traction as they solve the memory bottleneck of traditional von Neumann architectures~\cite{isaac}. 
While PIM architectures apply to different types of memories, including SRAM and eDRAM, it is especially advantageous to use embedded non-volatile memories (eNVM), due to their higher storage density and multi-level-cell (MLC) capability \cite{ReRAM,PCM,CMOS-MLC}. Moreover, their non-volatility and power efficiency are especially well suited for inference tasks that require relatively fixed NN parameters. PIM with eNVMs not only avoids high-energy, long-latency, off-chip DRAM accesses by densely storing all NN parameters on-chip, it also minimizes inefficient on-chip data movement 
and intermediate data generation by embedding critical multiplication and accumulation (MAC) computations within the memory arrays.

The resulting highly-efficient MAC computations within the memory arrays are, however, analog in nature, which raises some issues. First, analog computations are sensitive to noise from the memory devices and circuits. Luckily, inherent noise resilience of NN models ought to ameliorate this concern. 
Second and more importantly, typical NNs use high-precision neuron activations that require DACs and ADCs in order to feed inputs to and resolve MAC outputs from the memory arrays, respectively. These circuit blocks introduce significant area, power, and latency overheads. For example, the die photo in \cite{Naveen-BL-DAC} shows the area of 5-bit input DACs consume more than one third of the area of the SRAM PIM array. This translates to even greater relative overhead for an eNVM-based PIM. Thorough design space exploration of a ReRAM PIM architecture in \cite{isaac} shows the optimal configuration is to share one 8-bit ADC across all 128 columns of a 128x128 memristor array. Yet, this single ADC still occupies 48 times the area and consumes 6.7 times the energy of the entire memristor array. In fact, given this high cost of ADCs and DACs, many PIM designs resort to feeding and/or resolving activations one bit at a time in a sequential fashion, which then results in large latency penalties \cite{isaac,pipelayer-serial-AD-DA}. Either way, the high overhead introduced by ADCs and DACs defeats the original goal of attaining speed, power, and area efficiency by using PIM, which has recently drawn wide attention across the device, circuit, and algorithm communities \cite{EETimes,2020ANN,2020Review}. 
Thirdly, weights of conventional NNs usually require higher resolution (e.g., 8 bits) than are available in typical MLCs (e.g., 2 or 3 bits), which requires each weight to be across multiple memory cells and further degrades area efficiency. 

Reducing bit precision of both activations and weights can mitigate DAC and ADC overhead and reduce the number of cells needed for each weight. 
Hence, some PIM designs implement binary neural networks (BNN), with both binary weights and binary activations, obviating DACs and ADCs entirely, and only needing one single-level cell (SLC) per weight \cite{PIM-BNN-SenseAmp}. 
However, BNNs are not optimal for two reasons. First, the most popular BNN training algorithms use the straight-through estimator (STE) to get around BNN's indifferentiability problem during backpropagation \cite{Bengio2013,QNN,XNOR-net}. However, as we will show in Sections~\ref{TWD} and \ref{MNIST}, STE for binarizing activations is effective for training feedforward NNs, such as fully-connected (FC) and convolutional NNs (CNN), but works poorly for recurrent NNs (RNN). Second, binary weights in BNNs are too stringent to maintain high inference accuracy and do not take full advantage of the MLC capabilities of eNVMs.

\textcolor{\ReRevColor}{
	This paper proposes an ADC-/DAC-free PIM architecture using dense MLC eNVMs (Figure 1(c)), which addresses all of the aforementioned issues of prior PIM designs. The major contributions of this work are:}
\begin{itemize}[noitemsep,topsep=0pt] 
    \item \textcolor{\ReRevColor}{
		    To enable this optimal PIM architecture, we present} a new {\it noisy neuron annealing} (NNA) algorithm to train NNs with binary activations (BA) and multi-level weights (MLW) that take full advantage of dense MLCs. This algorithm achieves higher inference accuracy than using STE to train BA-MLW RNNs. 
    \item \textcolor{\RevisionColor}{We design an ADC-/DAC-free trigger word detection PIM accelerator with MLC eNVM, using a BA-MLW gated recurrent unit (GRU). 
Simulation results demonstrate superior inference accuracy and noise resilience compared to alternative algorithms.} 
    \item \textcolor{\RevisionColor}{
Using detailed hardware noise models and circuit co-design techniques, we validate our NNA training algorithm yields high inference accuracy and robustness against a variety of real hardware noise sources.}
    \item We further demonstrate the generality of our NNA algorithm by also applying it to feedforward networks. 
\end{itemize}
\section{Background}
\textbf{\textit{MLC eNVM for PIM.}} Before we introduce the PIM architecture, it is important to first review the technologies available for its critical building block -- the memory array. Although the memory array can be built with conventional SRAM cells, it is more advantageous to use eNVM, including traditional embedded Flash (eFlash) \cite{SONOS-PIM}, or emerging resistive RAM (ReRAM) \cite{ReRAM} and phase change memory (PCM) \cite{PCM}, or more recently, the purely-CMOS MLC eNVM (CMOS-MLC) \cite{CMOS-MLC}. 
Compared with SRAM, which is inherently binary (single level cells, SLC), eNVM's SLCs offer much higher area efficiency. Moreover, eNVM is often analog in nature that enables MLC capability for even higher storage density. Programming eNVM typically involves a continuous change in the conductivity of the memory devices, enabling them to store multiple levels of transister channel current in the cases of eFlash and CMOS-MLC, or multiple levels of resistor conductivity in the cases of ReRAM and PCM. The programming speed of eNVM is much slower than SRAM, but NN parameters are typically written infrequently and held constant during inference, rendering programming speed non-critical for inference-only applications. In fact, eNVM's non-volatility offers energy savings and obviates  reloading weights at power-up.

\textbf{\textit{PIM for NN inference.}} A conventional PIM architecture for NN inference is shown in Figure 1a \cite{isaac}. The weight matrix of a NN is directly mapped into the memory array 
and, because weights for NNs typically require high resolution (e.g., $\geq 8$ bits), multiple lower-resolution memory cells are often combined to represent one weight. 
This PIM structure can perform a matrix-vector multiplication in one step. Each input activation (i.e., $X_i, X_{i+1}, ...$) is simultaneously fed into individual wordlines as an analog voltage signal via a wordline DAC (WL-DAC), which then becomes a current through each memory cell proportional to the product of the input voltage and memory-cell conductance. MAC results are accumulated along corresponding sets of parallel bitlines (BLs) and resolved by column ADCs before being sent to digital nonlinear activation function units. A pair of columns are used to support both positive and negative weight values and each column pair corresponds to a single neuron.
Again, as discussed in Section~\ref{intro}, while these DACs and ADCs support high-resolution activation, they impose large area and power overheads.

\textbf{\textit{Quantization and BNN.}} Many different NN quantization algorithms have been proposed to reduce the bit widths of weights and/or activations while maximizing accuracy \cite{QNN,XNOR-net,Ternary-net}, in order to reduce storage and computation. For PIM, aggressive quantization can further relieve AD/DA resolution requirements for activations. In particular, BNNs with 1-bit activations and weights, can translate into much simpler PIM circuits, as shown in Figure 1b (compared to the conventional PIM architecture in Figure 1a). 
Since the activations are binary, WL-DACs and BL-ADCs in Figure 1a can be replaced by digital WL-drivers and conventional sense-amp comparators, respectively, both of which are compact peripheral components in standard memories. However, PIM implementations of BNNs have two major drawbacks. First, as shown in Figure 1b, 
binary weights use eNVM cells as 1-bit SLCs, not taking advantage of their MLC capability. 
Second, the most popular existing algorithm for training binary activations uses STE \cite{Bengio2013,QNN}, 
which is effective for feedforward NNs, but performs poorly on RNNs.

\section{Training BA-MLW NNs for optimal PIM implementation}
To avoid DACs and ADCs and fully leverage MLCs, we propose a BA-MLW NN structure, shown in Figure 1c, as an optimal design for PIM. For simplicity, memory devices are illustrated as resistor cells (omitting access transistors), corresponding to ReRAM or PCM, with multi-level weights encodes via their conductance values; other eNMV technologies, such as eFLASH and CMOS-MLC, directly encode weights into access transistor channel currents. 
The current difference across a pair of cells represent one postive or negative weight value. Moreover, binary activation (BA) only requires digital WL-drivers and sense-amps, obviating expensive DACs and ADCs.
In order to effectively train BA-MLW NNs, we propose a new algorithm that achieves high accuracy and resilience to quantization and noise for both feedforward and recurrent NNs.

\subsection{Training binary activations (BA)}
Binarizing the activations while maintaining high performance is challenging, because it not only restricts the expressive capacity of the neurons, but also introduces discrete computation nodes that preclude gradient propagation during training. We first review the STE algorithm prior to introducing our proposed BA training algorithm.

\textbf{\textit{Reviewing STE.}} STE applies to a stochastic binary neuron (SBN) \cite{Bengio2013}. During forward propagation of training, each neuron generates a binary output from a Bernoulli sample
\begin{equation}
\label{SBN}
	x_\text{SBN} =
	\begin{cases}
		1, \text{ with probability   } p = \mathrm{sigmoid}(s \cdot x)   \\
		0, \text{ with probability   } 1 - p \\
	\end{cases}
\end{equation}
in which $x$ is a pre-activation from the linear MAC and the logistic $\mathrm{sigmoid}$ function has a tunable slope $s$ \cite{STE-slope-anneal}. 
The SBN function is discrete with random sampling and, thus, does not have a well-defined gradient. Hence, STE simply passes through the gradient of the continuous sigmoid function during backpropagation
\begin{equation}
\label{STE}
	\frac{\partial L}{\partial x} = \frac{\partial L}{\partial {x_\text{SBN}}} \cdot \mathrm{sigmoid}'(s \cdot x) 
\end{equation}
in which $L$ is the loss function. In other words, it ignores the random discrete sampling process, and pretends the forward propagation implements a sigmoid function. The issue with STE is that propagating gradients w.r.t.\ the sample-independent mean ($\overline{x_\text{SBN}}=\mathrm{sigmoid}(s \cdot x)$) while ignoring the random sampling outcome can cause discrepancies between the forward and backward passes \cite{Gumbel-Eric}. 
In fact, STE is a biased estimator of the expected gradient, which cannot even guarantee the correct sign when back-propagating through multiple hidden layers \cite{Bengio2013}. Nonetheless, 
STE has been found to work better in practice than other more complicated gradient estimators for feedforward NNs \cite{Bengio2013}, which we also verify in Section~\ref{MNIST}. However, as shown in Section~\ref{TWD}, STE performs poorly when training RNNs with BAs.

\textbf{\textit{Proposed noisy neuron annealing (NNA) algorithm.}} 
We use the following noisy continuous neuron (NCN) function during the forward pass of training
\begin{equation}
\label{NCN_forward}
	x_\text{NCN} = \mathrm{sigmoid}(\frac{x+n_\text{train}}{\tau})
\end{equation}
in which we add an i.i.d.\ zero-mean Gaussian random variable (RV) $n_\text{train}\sim N(0,\sigma_\text{train}^2)$ to each pre-activation 
before passing into a continuous sigmoid function with temperature $\tau$. Equation~\ref{NCN_forward} can be broken down into two steps: (i) a noise injection step, $\tilde{x}=x+n_\text{train}$, and (ii) a continuous relaxing step, $x_\text{NCN}=\mathrm{sigmoid}(\tilde{x}/\tau)$. This noise injection step -- random noise added into pre-activations -- corresponds to quantization noise, due to binarizing activations and quantizing weights, that flows forward through the MAC. Therefore, if we train the NN with noise explicitly added into pre-activations, the NN would develop resilience to these quantization errors. The continuous relaxing step is inspired by the Gumbel-softmax trick \cite{Gumbel-Eric,Gumbel-Maddison}, which uses a sharpened sigmoid to approximate the binary step function while still allowing smooth gradients to flow. 
As Section~\ref{TWD} will show, 
it is important to start from a large value of the hyperparameter $\sigma_\text{train}$ to begin training with large noise and then anneal down to a smaller value, hence, the name of our training strategy -- ``noisy neuron annealing'' (NNA) algorithm.

Related to the additive noise in variational autoencoders, the Gaussian noise distribution complies to the ``mean and variance'' form required by the re-parameterization trick \cite{Gumbel-Maddison}. This has a nice Gaussian gradient identity property \cite{VAE-Rezende} that allows reversing the order between taking the expectation and taking the derivative. 
Combined with the continuous relaxation step, backpropagation through the entire NCN function does not encounter any discrete or sampling nodes:
\begin{equation}
\label{NCN_backward}
	\frac{\partial L}{\partial x} = \frac{\partial L}{\partial {x_\text{NCN}}} \cdot \mathrm{sigmoid}'(\frac{x+n_\text{train}}{\tau})
\end{equation}

During inference, we use the following noisy binary neuron (NBN) function:
\begin{equation}
\label{NBN}
	x_\text{NBN} = 
	\begin{cases}
		1, \text{ if } x+n_\text{eval} > 0   \\
		0, \text{ otherwise} \\
	\end{cases}
\end{equation}
which also has an additive i.i.d.\ Gaussian RV $n_\text{eval}\sim N(0,\sigma_\text{eval}^2)$, but replaces the continuous sigmoid in NCN with a discrete step function. In Sections~\ref{TWD} and \ref{MNIST}, we evaluate the noise resilience of trained NNs by sweeping $\sigma_\text{eval}$.

Prior work has studied the regularization effect of noise injection regarding its impact on NN generalization and noise resilience \cite{1994-MLP-noise,1996-RNN-noise,1996-noise-generalization,1995-similarities}. They use Taylor expansion of the loss function to show that adding Gaussian noise is akin to adding an extra regularization penalty term to the original loss function $L$, such that the effective loss becomes
\begin{equation}
\label{Loss_penalty}
	\tilde{L} = L+P = L + \frac{1}{2}\sigma_\text{train}^2\sum_{i}(\frac{\partial{L}}{\partial{x_i}})^2 
\end{equation}
where $x_i$ refers to a certain noise-injected node. In our case, $x_i$ includes all pre-activations. The regularization term, $P$, penalizes large gradients of $L$ w.r.t.\ noise-injected nodes, encouraging these nodes to find ``flatter regions'' of the solution space that are less sensitive to noise perturbations. Hyperparameter $\sigma_\text{train}$ controls the tradeoff between reducing the raw error $L$ and enhancing noise resilience.  
Specifically, for NCN activations to counteract the noise, they tend to give up the highly-expressive but noise-prone transition region of the sigmoid and, instead, develop a bimodal pre-activation distribution to push them into the saturated regions, close to 1 or 0, that are highly immune to noise \cite{Semantic-Hashing}.
Equation \ref{Loss_penalty} also provides a quantitative metric to estimate the noise resilience a NN acquires during training. We derive a detailed form to calculate this penalty term in Section~\ref{TWD} to compare amongst different NNs.

\begin{table}[h]
	\caption{An example mapping 7-level weights into the $I_\text{cell}^+$ and $I_\text{cell}^-$ current magnitudes of a pair of 4-level cells. $I_\text{fs}$ is the full-scale current that corresponds to $\alpha$.}
  \label{MLW}
  
\vspace*{2ex}
  \begin{tabular}{cccccccc}
    \toprule
	  weight&$-\alpha$&$-\frac23\alpha$&$-\frac13\alpha$&$0$&$\frac13\alpha$&$\frac23\alpha$&$\alpha$\\
    \midrule
	  $I_\text{cell}^-$&$I_\text{fs}$&$\frac23I_\text{fs}$&$\frac13I_\text{fs}$&$0$&$0$&$0$&$0$\\
    \midrule      
	  $I_\text{cell}^+$&$0$&$0$&$0$&$0$&$\frac13I_\text{fs}$&$\frac23I_\text{fs}$&$I_\text{fs}$\\
  \bottomrule
  \vspace*{-8mm}
\end{tabular}
\end{table}

\begin{figure*}[t]
  \centering
	\includegraphics[width=0.9\linewidth]{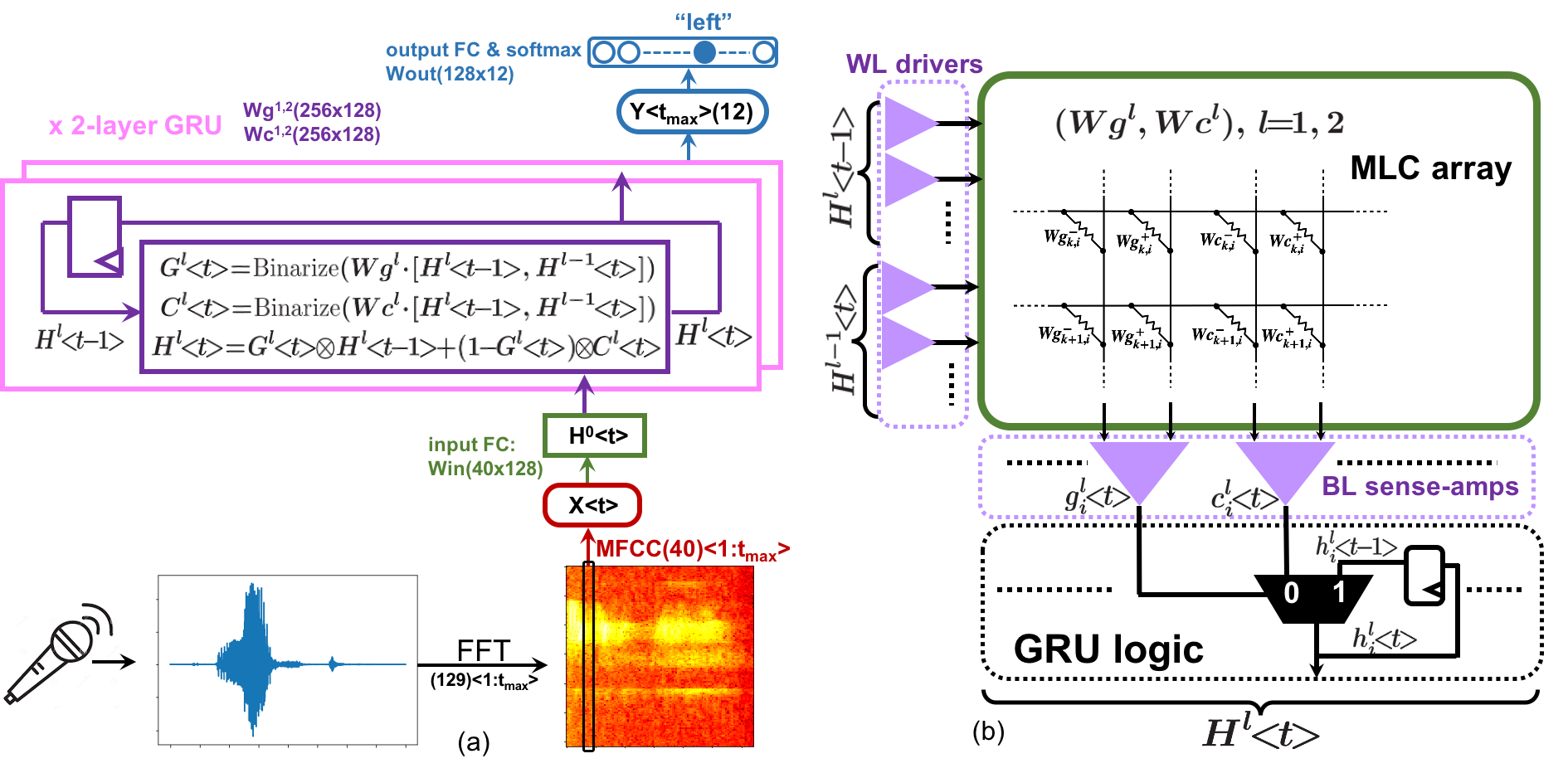}
	\caption{(a) BA-MLW GRU architecture with an input FC and 2 GRU layers and (b) the PIM implementation of a GRU layer.} 
     \vspace{-4mm}
  \label{TWD-Circuits}
\end{figure*}

\subsection{Training multi-level weights (MLW)}
Our NNA algorithm not only endows the NN with high resilience to binarizing activations, but also enables MLWs to leverage dense MLCs with high inference accuracy.
Each weight can be quantized down to a small number of levels capable of encoding with one pair of MLCs (Figure 1c), as opposed to needing to combine multiple memory cells for high-resolution weights (Figure 1a).
%
%

To quantize MLWs from full-precision (FP) weights, we first determine a suitable clipping range $[-\alpha, \alpha]$ for each weight matrix based on the weight distribution statistics from pre-training with FP weights. Then during fine-tuning with weight quantizations, we clip each weight matrix into $[-\alpha, \alpha]$ prior to quantizing the weights into evenly-spaced levels within this range. We follow the same practice as \cite{QNN} for training, i.e., we use the quantized weights in the forward pass, but still keep the FP weights and accumulate gradients onto FP weights in the backward pass. After training is complete, the FP weights can be discarded and only the quantized weights are used for inference. For the special cases of 3- and 2-level weights, we use the training algorithm in \cite{Ternary-net} for 3-level (ternary) and \cite{XNOR-net} for 2-level (binary) weights. 


Table~\ref{MLW} shows how to represent a 7-level weight via the current differential across a pair of 4-level (2-bit) MLCs. 
Following the same principle, 
a 15-level weight can be encoded with a pair of 8-level (3-bit) MLCs, while a 3-level weight can use a pair of binary cells (1-bit, SLC).

\section{\textcolor{\ReRevColor}{Application case studies}}
\textcolor{\ReRevColor}{We present two case studies that apply our NNA algorithm to (1) an RNN for a trigger word detection task and (2) a feedforward NN for handwritten digit recognition. We focus on the first case study to thoroughly demonstrate the merits of a BA-MLW RNN trained using the NNA algorithm. The second case study confirms the algorithm further generalizes to feedforward NNs.
%
}
\vspace*{-1ex}
\subsection{A trigger word detection PIM accelerator using BA-MLW GRU with MLC eNVM}
\label{TWD}
Trigger word detection is an important always-ON task for speech-activated edge devices, for which power and cost efficiency is paramount. We use the {\it Speech Commands} dataset from \cite{Speech-Commands-Warden}, which consists of over 105,000 audio clips of various words uttered by thousands of different people, with a total of 12 classification categories: 10 designated keywords, silence, and unknown words.
\textcolor{\ReRevColor}{
	We first present the architectural designs of the NN and PIM accelerator (Section \ref{arch}), then elaborate on the software training results using different training methods (Section \ref{software_train}), and finally evaluate expected hardware performance with detailed noise models (Section \ref{noise-model}).}
	
\subsubsection{\textcolor{\ReRevColor}{The architecture of the NN and PIM accelerator}}
\label{arch}
\textbf{\textit{The NN model structure.}} RNNs are well-suited to this speech-recognition task. Figure 2a illustrates a 2-layer gated recurrent unit (GRU) \cite{GRU-paper} with BA and MLW.    
We first perform FFT (window=16ms, stride=8ms) on the raw audio signals and then extract 40 Mel-frequency cepstral coefficients (MFCC) per 8ms timestep. Each MFCC vector passes through an input FC layer that encodes it into a 128 dimensional binary vector as the input to the first layer of a 2-layer stacked GRU (both layers use 128 dimensional vectors). 
We use the following modified version of GRU equations \cite{Andrew-Ng-GRU}:
\begin{align}
\label{GRU_equation}
	    \widetilde G^l<t> &= Wg^l  \cdot  [H^l<t-1>, H^{l-1}<t>]\\
    	G^l<t>&=f(\widetilde G^l<t>)\\
    	\widetilde C^l<t> &= Wc^l  \cdot  [H^l<t-1>, H^{l-1}<t>]\\
        C^l<t>&=f(\widetilde C^l<t>)\\
	    H^l<t>&= G^l<t>\otimes H^l<t-1>+(1-G^l<t>)\otimes C^l<t> \label{mux}
\end{align}
%
where $t$ denotes the timestep, $l$ is the layer number, and the gate $G^l<t>$ ($l=1,2$), candidate $C^l<t>$ ($l=1,2$), hidden state $H^l<t>$ ($l=1,2$), and the input encoding $H^0<t>$ are all 128 dimensional activation vectors, trained using our NNA algorithm. 
The activation function $f$ refers to NCN (Equation \ref{NCN_forward}) during training, and NBN (Equation \ref{NBN}) for evaluation, and simply uses the binary step function for PIM deployment (Figure 2a and 2b). 
Compared with the original GRU equations from \cite{GRU-paper}, we remove the reset gate since it has minimal effect on the accuracy of this task, but greatly simplifes circuit design (Figure 2b). 
After the GRU processes inputs from all timesteps, the final timestep output of the top layer feeds into an output FC layer followed by a 12-way softmax that yields the classification. 

\textbf{\textit{PIM circuit design for the GRU.}} The BA-MLW GRU equations can be mapped into compact PIM circuits shown in Figure 2b. 
The MLC array implements MAC computations and the column sense-amps resolve the binarized $G^l<t>$ and $C^l<t>$. $G^l<t>$ serves as the multiplexer selection signal (since $1-G^l<t>=!G^l<t>$ in Equation~\ref{mux} for binary signals) to choose either to keep the binary hidden state saved from the previous timestep or to update it with the current binary candidate state. 
Since all analog MAC signals are encapsulated inside the MLC array, all input/output interface signals are binary, and the GRU logic outside the array is totally digital, {\bf this PIM design does not require any ADCs or DACs}.

\begin{figure*}[h]
  \centering
  \includegraphics[width=\linewidth]{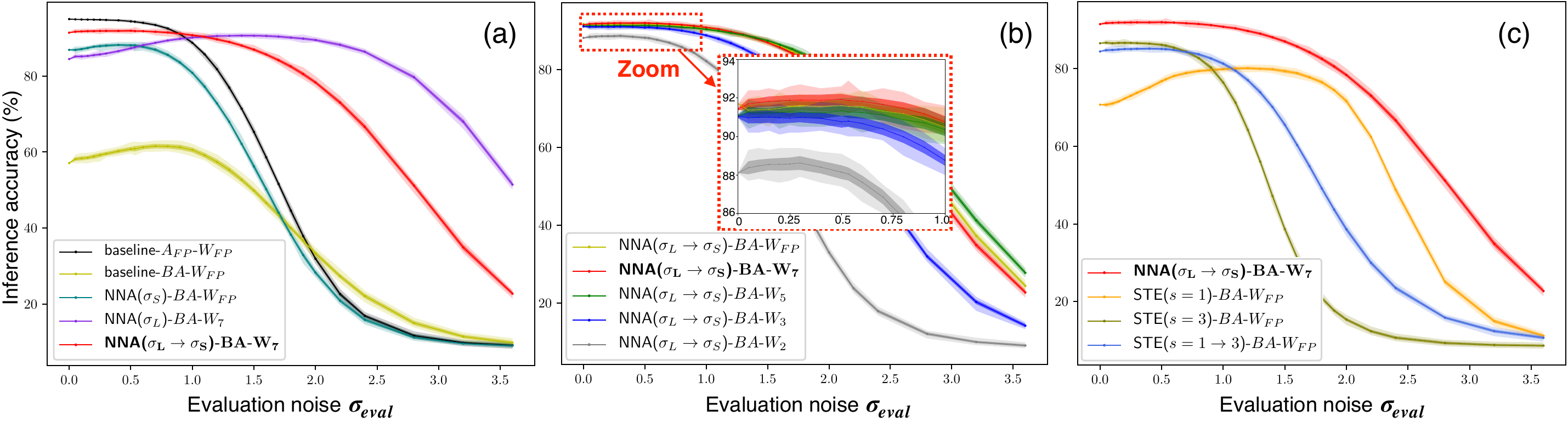}
     \vspace{-6mm}
	\caption{Inference accuracy vs $\sigma_{eval}$ of evaluation noise of GRUs trained with different methods (plotting the mean accuracy surrounded with the ranges of mean$\pm$STD and max/min). 
	(a) compares 4 networks trained through different stages, in which the baseline network is evaluated with both FP activations and BAs, (b) compares different weight quantization levels, using NNA algorithm, and (c) compares NNA with STE algorithm. Except for baseline-$A_{FP}$-$W_{FP}$ that uses FP activations, all the other curves are evaluated with BAs using NBN Equation \ref{NBN}.}
     \vspace{-3mm}
  \label{GRU_acc_plots}
\end{figure*}

\textbf{\textit{Noise-induced loss penalty terms for GRU.}} 
For the GRU, we can derive the noise-induced regularization penalty term $P$ in Equation~\ref{Loss_penalty} by taking the derivatives of loss $L$ w.r.t.\ the pre-activations of candidates ($\widetilde C^l<t>$) and gates ($\widetilde G^l<t>$).
\begin{align}
	P &= \frac{1}{2}\sigma_{train}^2\sum_{i,t,l}(\frac{\partial{L}}{\partial{h_i^l<t>}})^2\cdot \{Pg+Pc\} \label{Loss_penalty_GRU} 
\end{align}
\begin{align}
	Pg &= [(h_i^l<t-1>-c_i^l<t>)\cdot \mathrm{sigmoid}'(\widetilde g_i^l<t>)]^2 \label{Pg}\\
	Pc &= [(1-g_i^l<t>)\cdot \mathrm{sigmoid}'(\widetilde c_i^l<t>)]^2 \label{Pc}
\end{align}
where a lowercase letter with subscript $i=0\sim 127$ denotes one element of the corresponding uppercase vector, and $Pg$ and $Pc$ regularize gates and candidates, respectively.
%
The forms of Equations \ref{Pg} and \ref{Pc} have intuitive interpretations when minimizing them during training. To minimize $Pg$, one way is to reduce the derivative of the gate ($\mathrm{sigmoid}'(\widetilde g_i^l<t>)$) by pushing $\widetilde g_i^l<t>$ away from zero (the steep slope region of sigmoid), so that the gate is firmly ON or firmly OFF; alternatively, it can try to make the candidate of the current timestep $c_i^l<t>$ equal to the hidden state of the previous timestep $h_i^l<t-1>$, such that the new hidden state $h_i^l<t>$ would be the same regardless of $\widetilde g_i^l<t>$. Either way, minimizing $Pg$ makes $h_i^l<t>$ immune to noise injected into $\widetilde g_i^l<t>$. To minimize $Pc$, training will either reduce the gradient of candidate ($\mathrm{sigmoid}'(\widetilde c_i^l<t>)$), by pushing $\widetilde c_i^l<t>$ into the saturated flat regions of sigmoid, or try to turn on the gate $g_i^l<t>$ to preserve the hidden state from the previous timestep $h_i^l<t-1>$ disregarding the new candidate $c_i^l<t>$. Either way, it desensitizes $h_i^l<t>$ to noise injected into $\widetilde c_i^l<t>$. 
Equations \ref{Loss_penalty_GRU}--\ref{Pc} provide a quantitative metric to assess the resilience to quantization and noise in a trained GRU, which we use to compare different training schemes.

\subsubsection{\textcolor{\ReRevColor}{Software training results}}
\label{software_train}
\textcolor{\ReRevColor}{To quantitatively compare NN accuracy and noise-resilience trained with different methods, we inject the same type of Gaussian training noise during inference and sweep the noise sigma -- a common practice adopted in previous NN noise-resilience studies \cite{2019IEDM,2019-UCSB-IJCNN-ShotNoise,2019-DAC,2019-TVLSI-SAR-VOS,2020-IRPS}. We first compare different methods of binarizing activations, then explore different quantization levels for the weights, and, finally, compare RNN training performance using our NNA algorithm versus the popular STE algorithm. Results are summarized in Figure \ref{GRU_acc_plots}.}

\textbf{\textit{\textcolor{\ReRevColor}{Comparing training schemes for binarizing activations.}}} As a full-precision (FP) baseline, we first train the NN with FP sigmoid activations and FP weights, without noise injection or quantization. During inference, we add noise $n_\text{eval}\sim N(0,\sigma_\text{eval}^2)$ to its pre-activations and evaluate it with both FP sigmoid activations (baseline-$A_\text{FP}$-$W_\text{FP}$, black), and binarized activations (baseline-$BA$-$W_\text{FP}$, yellow), 
shown in Figure 3a. 
Trained without noise injection, baseline-$A_\text{FP}$-$W_\text{FP}$ 
cannot maintain its high accuracy at large $\sigma_\text{eval}$, making it vulnerable to quantization errors, leading to the poor performance of baseline-$BA$-$W_\text{FP}$. 


To endow the NN with resilience to quantization errors, we use our NNA algorithm and retrain from the FP sigmoid baseline (which is a good initialization point to speed up retraining). Initially, we use large training noise $\sigma_\text{train}=\sigma_L=1.6$, and plot the inference accuracy with BAs and 7-level weights (NNA($\sigma_L$)-$BA$-$W_7$, purple) in Figure 3a. Results show high accuracy across a wider range of $\sigma_\text{eval}$ than baseline-$BA$-$W_\text{FP}$, demonstrating stronger resilience to quantization errors. However, the accuracy peaks at a large $\sigma_\text{eval}$ around $\sigma_L$, and lower at small $\sigma_\text{eval}$, because it is trained to minimize its loss in the presence of this large additive noise. This noise-resilience profile might suit certain noisy circuit environments, e.g., noisy power supply. 

In order to also achieve high accuracy at small $\sigma_\text{eval}$, we anneal $\sigma_\text{train}$ down to a small $\sigma_\text{train}=\sigma_S$ and further retrain. As shown in Figure 3a (NNA($\sigma_L\rightarrow \sigma_S$)-$BA$-$W_7$, red), peak accuracy is achieved at smaller $\sigma_\text{eval}$, demonstrating the efficacy of NNA with annealing (from $\sigma_L$ to $\sigma_S$). 
In comparison, directly retraining with $\sigma_S$ from the baseline (NNA($\sigma_S$)-$BA$-$W_7$, green), without the intermediate $\sigma_L$ stage, results in much worse accuracy and noise tolerance than NNA with annealing. 

In practice, we find the choices of the hyperparameter $\sigma_\text{train}$ quite flexible.  
For the initial large noise, we use a $\sigma_L$ to be about 20\% of the standard deviation (STD) of the inherent pre-activation distribution, corresponding to $\sigma_L=1.6$ for this GRU, though a wide range of values all work similarly well; for the annealed noise, we find $\sigma_S=0\sim 0.5$ all achieves optimal results.  
For temperature $\tau$ in NCN, we find $0.3$ to be optimal: If too small, RNN's gradient explodes. If too large, BA is not sufficiently approximated.

\begin{table}
  \vspace{-0mm}
  \caption{Loss penalty terms of the 4 NNs in Figure 3a.}
  \label{penalty_terms}
	\resizebox{0.48\textwidth}{!}{
  \begin{tabular}{ccccc}
    \toprule
	  Network&FP baseline&NNA($\sigma_L$)&NNA($\sigma_L\rightarrow \sigma_S$)&NNA($\sigma_S$)\\
    \midrule
	  Pg+Pc&1.0000&0.5034&0.5018&0.8272\\
  \bottomrule
  \vspace{-8mm}
  \end{tabular}}
\end{table}

\textbf{\textit{Loss penalty term interpretations.}} To understand why the annealing procedure of NNA is critical, we compare the loss penalty terms ($Pg+Pc$ from Equation \ref{Pg} and \ref{Pc}) of the four networks in Figure 3a. Shown in Table \ref{penalty_terms}, we normalize them by the penalty value of baseline-$A_\text{FP}$-$W_\text{FP}$, since only the relative values matter for comparison. After retraining with $\sigma_L$, NNA($\sigma_L$)-$BA$-$W_7$'s penalty term reduces to half of the baseline penalty, explaining its higher resilience to noise and quantization errors. After further retraining with annealed $\sigma_S$, NNA($\sigma_L\rightarrow \sigma_S$)-$BA$-$W_7$ maintains this small penalty value, even though $\sigma_S$ provides less regularization effect -- the smaller multiplier $\sigma_S^2$ in Equation \ref{Loss_penalty_GRU} (compared to previous $\sigma_L^2$) makes the retraining prioritize reducing the raw error (thus higher peak accuracy) over enhancing noise resilience.  
This means the network can still ``memorize'' its previous large-noise training regularization effect even after annealing to fine-tune with smaller noise. In contrast, without the intermediate ``experience'' of large noise training, NNA($\sigma_S$)-$BA$-$W_7$ has much less regularization effect to reduce its loss penalty.   

\textcolor{\arXivColor}{It should also be pointed out that if evaluated with FP activations and zero noise, all these networks achieve similarly high accuracy as the purely FP sigmoid network, but their accuracy and noise resilience are dramatically different after using binary activations. This implies that trained through different stages, the 4 networks in Figure 3a find distinct regions in the global solution space: retraining with $\sigma_L$ finds a promising solution region that's insensitive to quantization errors, while the further fine-tuning with $\sigma_S$ only does a local search to optimize accuracy at the small noise range; in contrast, training without noise injection or only with small noise will not discover the solution region that's resilient to noise and quantization due to lack of regularization penalty.}

\textbf{\textit{Weight quantization levels.}} Figure 3b shows the GRU's performance across different numbers of weight quantization levels and confirms the high resilience to weight quantization offered by our NNA algorithm. The resulting performance of 7-level (implemented with a pair of 4-level cells) and 5-level (a pair of 3-level cells) quantization are almost the same as using FP weights. GRUs with 7-level weights achieve the highest peak accuracy at the lower end of $\sigma_\text{eval}$ range. Accuracy for GRUs with ternary weights (a pair of 2-level cells) degrades more quickly as $\sigma_\text{eval}$ increases. In contrast, GRUs with binary weights, which can be implemented with SRAMs), suffer the most accuracy degradation across the entire $\sigma_\text{eval}$ range. Our NNA algorithm's high resilience to weight quantization makes it possible to use a pair of 3 or 4-level MLCs to achieve performance comparable to GRUs with FP weights. Even GRUs with tenary weights can achieve relatively high inference accuracy for applications with small $\sigma_\text{eval}$. Therefore, our optimal PIM architecture (Figure 1c) avoids combining multiple cells to achieve high-resolution weights (Figure 1a, \cite{isaac}), thereby saving area and simplifying circuit design.

\textbf{\textit{NNA vs STE.}} We also experiment with SBN trained with STE, and compare its performance with our NNA algorithm in Figure 3c. We try 3 different settings for slope $s$: $s=1$ (yellow), $s=3$ (blue), and annealing $s$ from $1$ to $3$ ($s=1\rightarrow 3$, green) \cite{STE-slope-anneal}. Even with FP weights, all three GRUs trained with STE achieve worse inference accuracy compared to using NNA (NNA($\sigma_L\rightarrow \sigma_S$)-$BA$-$W_7$, red). 

\textcolor{\arXivColor}{We should also point out an important distinction between the forms of SBN (Equation \ref{SBN}) and our NCN (Equation \ref{NCN_forward}): SBN only has one parameter $s$, whereas NCN has two degrees of freedom using $\tau$ and $\sigma_{train}$. On one hand, SBN uses $s$ to control the sigmoid slope (corresponding to $1/\tau$ in NCN), and similar to $\tau$, we find $s$ needs to be no greater than about 3, in order not to run into exploding gradient problem. On the other hand, $s$ also controls the stochasticity of the Bernoulli sampling: a smaller $s$ introduces more randomness thus a higher noise resilience range, as can be seen from Figure 3c, comparing $s=1$, $s=3$ and $s=1\rightarrow 3$. However, SBN cannot seperately control the the sigmoid slope and the stochasticity. In contrast, our NCN has independent controls: $\tau$ is chosen to approximate binary outputs while avoiding exploding gradients, whereas the magnitude of noise injection is seperately controlled by $\sigma_{train}$. This flexiblity enables us to effectively implement NNA algorithm's annealing procedure using NCN.} 

\textcolor{\arXivColor}{From a mathematical rigor point of view, in contrast to the Gaussian RVs used in NCN, the Bernoulli RVs used in SBN do not comply with the ``location-scale'' distribution required for using the reparameterization trick \cite{Gumbel-Maddison}. Therefore, it is mathematically illegal for STE to change the order between taking expectation and taking derivative for Bernoulli RVs (Equation \ref{STE}). 
Increasing the slope $s$ can alleviate the discrepancy between the forward and backward pass of SBN (making the math less wrong, which explains the higher accuracy with $s=3$ in Figure 3c), but due to the lack of separate controls, changing $s$ inevitably changes both the sampling randomness and sigmoid's gradient, and $s$ cannot be too large which will cause exploding gradients.}

\subsubsection{\textcolor{\RevisionColor}{Hardware noise model and validation results\\}}
\label{noise-model}
\textbf{\textit{Need for accurate hardware noise models. }}
\textcolor{\RevisionColor}{So far we have used the same type of additive random Gaussian noise for both training and inference as $N_\text{train}$ and $N_\text{eval}$ in Figure~\ref{GRU_acc_plots}, consistent with prior work that 
evaluate and compare NN noise resilience 
\cite{2019IEDM,2019-UCSB-IJCNN-ShotNoise,2019-DAC,2019-TVLSI-SAR-VOS,2020-IRPS}. }	
\textcolor{\RevisionColor}{
While prior hardware noise modeling work generally over-simplify by lumping all the noise sources into a single random Gaussian RV, real circuits have a variety of noise sources with properties different from these additive Gaussian RVs, mandating realistic models to faithfully evaluate their impact on hardware inference accuracy.} 

{To evaluate how real circuit noise impacts inference accuracy, we account for the detailed profiles of three important sources of physical noise in PIM circuits shown in Figure~\ref{2021-noise-model}}\textcolor{\ReRevColor}{: (1) the weight noise $N_\text{MLC}$ due to variability of eNVM devices, (2) an offset error $N_\text{OS}$ due to device mismatch in the sensing circuitry, and (3) a white noise source $N_\text{white}$ from thermal and shot noise of the circuits. In the following subsections, we first elaborate on the distinctive characteristics of these three noise sources and present our simulation methodology of the overall noise model; then we introduce the devices and circuit designs of the PIM implementation used for deriving the statistics to build the noise model; finally, we present the hardware validation experiment results.}

\begin{figure}[h] 
  \centering
  \includegraphics[width=3.2in]{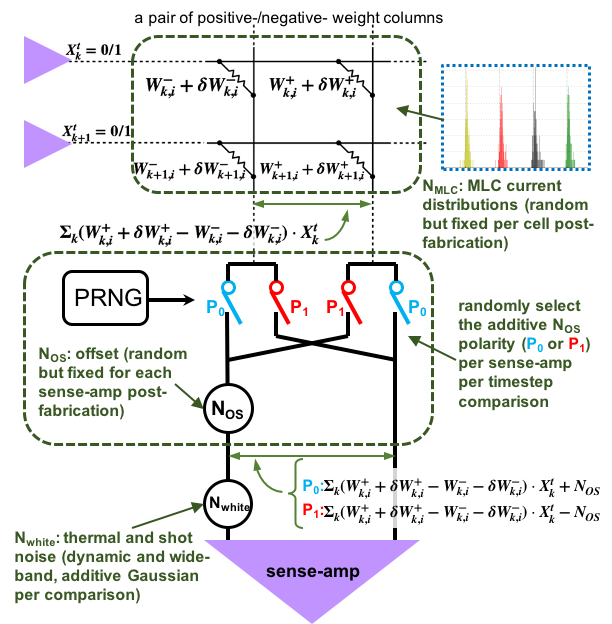}
	\caption{Illustration of detailed circuit-level noise sources in PIM hardware.}
  \label{2021-noise-model}
  \vspace{-1mm}
\end{figure}

\textbf{\textit{(1) $N_\text{MLC}$}} \textcolor{\RevisionColor}{comes from eNVM device manufacturing and programming variability, modeled as noise in programmed MLC weight levels, i.e., 
per-cell static weight error $\delta W_{k,i}^+$ or $\delta W_{k,i}^-$ deviating from the intended/ideal value $W_{k,i}^+$ or $W_{k,i}^-$ (Figure \ref{2021-noise-model}). Depending on input activations of each timestep, each cell's weight error selectively contributes noise to the total weight noise of $N_\text{MLC}=\Sigma_k (\delta W_{k,i}^{+}-\delta W_{k,i}^{-})\cdot X_k^t$ added to each corresponding element of MAC pre-activations. }

\textbf{\textit{(2) $N_\text{OS}$}} \textcolor{\RevisionColor}{models transistor-mismatch-induced offset as a sense-amp input-referred static error. Each sense-amp's random offset is determined after fabrication, which adds a fixed asymmetric bias term into the MAC computation associated with the comparisons of each sense-amp. This  offset proves to be detrimental to inference accuracy due to its lack of dynamic randomness, but most prior noise-modeling studies overlook this important source of error.}

Among the few papers that address this offset error, \cite{2019-TVLSI-SAR-VOS} simplifies it into a Gaussian RV, lumped with other noise sources by adding their variances together, and proposes per-chip, post-fabrication retraining; \cite{2018-ISSCC-SAR-VOS} uses on-chip digital calibration circuitry to cancel the offset after fabrication.
In contrast, we dedicate the realistic static noise model $N_\text{OS}$ to each offset error, and leverage a simple yet effective circuit technique that avoids the need for individual retraining or the overhead of offset calibration. 
Our solution is to randomly flip the polarity of the added $N_\text{OS}$ using a set of switches controlled by a pseudo random number generator (PRNG), as shown in Figure \ref{2021-noise-model}, 
such that each fixed $N_\text{OS}$ magnitude is presented to each element of pre-activations with a random polarity per sense-amp comparison per timestep, \textcolor{\ReRevColor}{thereby converting the effect of each sense-amp offset into a Bernoulli RV. } 
Compared with the smooth ``bell-shaped'' Gaussian noise with which our NNs are trained, Bernoulli distributions have very different forms, but they can guarantee per-element zero-mean and dynamic randomness across timesteps. These are the traits of noise that NNs trained with our NNA algorithm are surprisingly resilient to, as we will show in our validation experiments. 


\textbf{\textit{(3) $N_\text{white}$}} \textcolor{\RevisionColor}{represents thermal and shot noise from the circuits, which are dynamic random white noise, modeled as an input-referred Gaussian noise source at each sense-amp. This is the only truly random and dynamic source of hardware noise, with exactly the same properties as the Gaussian noise used for training.} 

\textbf{\textit{The overall noise model}} \textcolor{\RevisionColor}{accounts for the three aforementioned noise sources, such that the total noise added into each element of pre-activations at each timestep is: $N_\text{MLC} \pm N_\text{OS} + N_\text{white}= \Sigma_k (\delta W_{k,i}^{+}-\delta W_{k,i}^{-})\cdot X_k^t + (-1)^\text{polarity} \cdot N_\text{OS} + N_\text{white}$. When simulating the inference performance of a single chip, $N_\text{OS}$, $\delta W_{k,i}^+$ and $\delta W_{k,i}^-$ are determined/static after fabrication and weight programming, so we only randomly sample them once and fix them during inference on all validation examples; $N_\text{white}$ is dynamic and so we randomly generate a new sample for each sense-amp comparison; the PRNG dynamically generates a random selection of $polarity$ to add or subtract the fixed $N_\text{OS}$ magnitude for each corresponding sense-amp comparison. We validate hardware performance at the presence of all these noise sources by simulating cycle-accurate inference across GRU timesteps. } 

\textbf{\textit{Device models and circuit simulations.}} \textcolor{\RevisionColor}{For eNVM devices, we evaluate three promising emerging eNVMs (ReRAM, PCM, and CMOS-MLC) and model their corresponding $N_\text{MLC}$ based on measured data from \cite{ReRAM}, \cite{PCM}, and \cite{CMOS-MLC}, respectively. We design and simulate all peripheral circuits targeting a 16nm FinFET technology \cite{TSMC-16nm}. The sensing circuits that resolve BAs adopt the dynamic bitline discharge technique commonly used in memory reading circuitry, with the self-timed StrongArm-type sense-amp from \cite{SenseAmp}. Sensing via variable bitline discharge time automatically handles the wide statistical range of 
bitline currents resulting from PIM-based MAC computations. Transistor sizes in the sense-amp pose a tradeoff between area versus $N_\text{OS}$, i.e., enlarging the transistor sizes can reduce $N_\text{OS}$ at the expense of a larger sense-amp area overhead. We use Monte Carlo simulations to measure the $N_\text{OS}$ of a range of sense-amp sizes, and evaluate their impacts on inference accuracy. } 

\begin{table}
	\caption{Summary of hardware noise validation results.}
  \label{validation_table}
	\resizebox{0.48\textwidth}{!}{
  \begin{tabular}{cccc}
    \toprule
	  \begin{tabular}{@{}c@{}}Impact of $N_\text{OS}$\\(w/o $N_\text{MLC}$)\end{tabular}&
	  \begin{tabular}{@{}c@{}}$N_\text{white}$-only\\(baseline)\end{tabular}&
	  \begin{tabular}{@{}c@{}}$N_\text{white}+N_\text{OS}$\\fixed polarities\end{tabular}&
	  \begin{tabular}{@{}c@{}}$N_\text{white}+N_\text{OS}$\\flipping polarities\end{tabular}\\
    \midrule      
	  Accuracy range (\%)&$91.76\pm 0.23$&$91.26\pm 0.22$&$91.86\pm 0.24$\\
    \midrule \\
	\midrule      
	\begin{tabular}{@{}c@{}} Impact of eNVM \\(w/ all noise sources)\end{tabular}&
	  PCM&ReRAM&CMOS-MLC\\
    \midrule      
	  Accuracy range (\%)&$91.60\pm 0.34$&$91.71\pm 0.27$&$91.60\pm 0.29$\\
  \bottomrule
  \vspace{-6mm}
  \end{tabular}}
\end{table}

\begin{figure}[h] 
  \centering
  \includegraphics[width=0.8\linewidth]{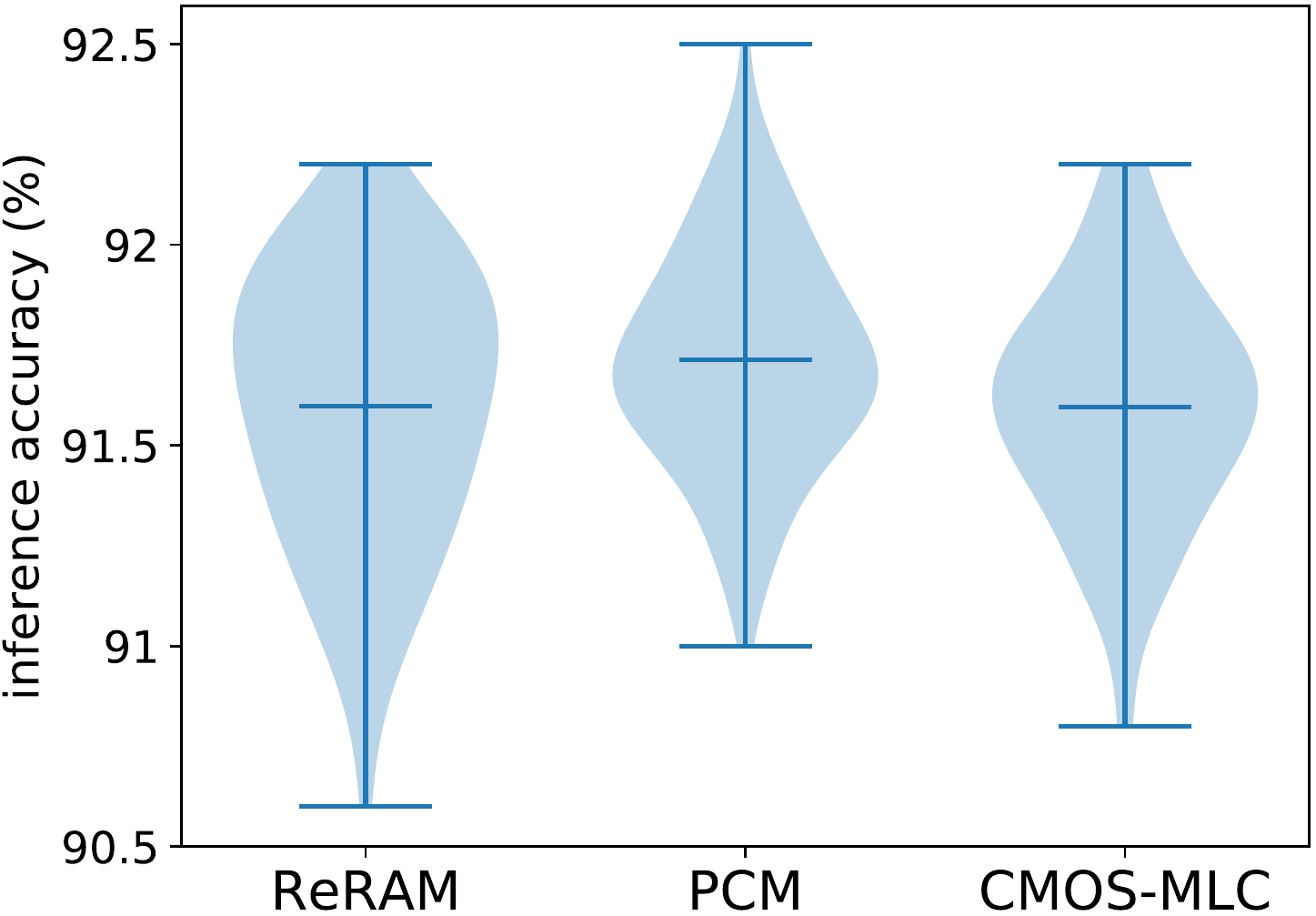}
	\caption{\textcolor{\ReRevColor}{ 
		Hardware inference accuracy distributions of the trigger word detection PIM accelerator using three promising eNVM technologies, validated with the detailed noise models in Figure~\ref{2021-noise-model}}.} 
  \label{hardware_noise_validation}
\end{figure}

\textbf{\textit{Validation results}} \textcolor{\ReRevColor}{ are summarized in Table~\ref{validation_table} and Figure~\ref{hardware_noise_validation}. We choose NNA($\sigma_L\rightarrow \sigma_S$)-$BA$-$W_7$ as the optimal design target for circuit implementation and evaluation of the NNA algorithm's resilience to hardware noise. We establish a baseline accuracy ($mean \pm \textit{STD} = (91.76\pm0.23)$\%) corresponding to when only $N_\text{white}$ is present and both $N_\text{OS}$ and $N_\text{MLC}$ are zero. This baseline accuracy is on par with the peak software inference accuracy in Figure~\ref{GRU_acc_plots}. The magnitude of $N_\text{white}$ for real circuits turns out to be too small to have measurable impacts on accuracy. We first focus on exploring the performance impact of $N_\text{OS}$ and the polarity flipping circuit technique. We then simulate with the entire noise model across three promising eNVM technologies.} 

\textcolor{\ReRevColor}{As summarized by the upper two rows of Table~\ref{validation_table}, when we inject the modeled $N_\text{OS}$ but fix their polarities, the validation accuracy 
drops to $mean\pm \textit{STD} = (91.26\pm0.22)$\%. 
However, by randomly flipping the polarity of each $N_\text{OS}$ using the PRNG, validation accuracy of the NN recovers 
to $mean\pm \textit{STD} = (91.86\pm0.24)$\%, proving the effectiveness of this simple random polarity-flipping technique. The modeled statistical magnitudes of $N_\text{OS}$ are derived from our sensing circuitry design with a reasonable sense-amp sizing choice that requires no more than 8 fins of minimum-length FinFETs for each input differential pair transistor. Therefore, our PIM accelerator design ensures minimal peripheral sensing circuitry overhead (mostly from sense-amps) thanks to BAs, in contrast to area- and power-consuming ADCs otherwise needed for higher-precision activations.} \textcolor{\RevisionColor}{These validation results also reveal that our NNA algorithm results in NNs that are resilient to noise profiles beyond the additive Gaussian noise with which they are trained (consistent with discoveries in \cite{2020-Nature,2016-TolerantVariety}) and they are particularly tolerant to dynamic random zero-mean noise, whereas the exact shape of the distribution is less important (no need to be smooth or bell-shaped). }

\textcolor{\ReRevColor}{Finally, we evaluate 
inference performance for PIM designs with binary activations and 7-level weights (each weight implemented with a pair of 2-bit (4-level) MLCs) using the three promising eNVM technologies (PCM, ReRAM, and CMOS-MLC) and with all three types of noise sources (plus random polarity flipping).
Figure~\ref{hardware_noise_validation} and the lower two rows of Table~\ref{validation_table} summarize the validation results. 
PCM, ReRAM, and CMOS-MLC achieve hardware inference accuracy of $mean\pm \textit{STD}$ = ($91.60\pm 0.34$)\%, ($91.71\pm 0.27$)\%, and ($91.60\pm 0.29$)\%, respectively, validating that PIM circuits with all three eNVM technologies can achieve performance comparable to the software accuracy (Figure 3) even in the presence of realistic device and circuit non-idealities.} 

\begin{figure*}[t]
  \centering
  \includegraphics[width=\linewidth]{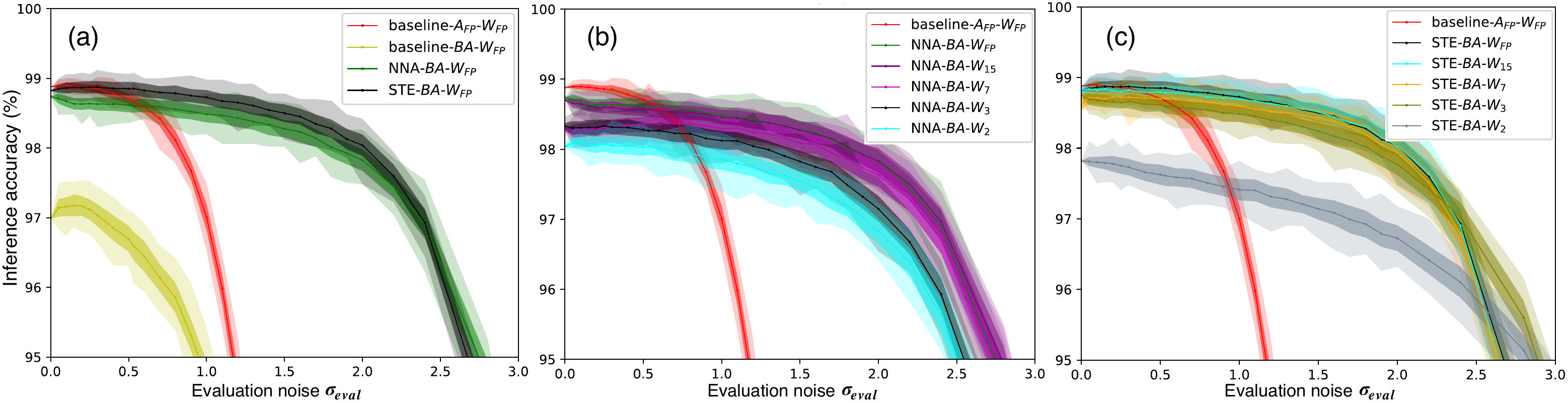} 
  \vspace*{-3ex}
	\caption{
	Comparisons of inference accuracy (showing $mean\pm \textit{STD}$ and max/min ranges
	) vs $\sigma_\text{eval}$ of evaluation noise 
	of LeNet5 trained with: (a) FP baseline (evaluated with both FP activations and BAs) vs NNA algorithm with BAs and STE algorithm with BAs, different weight quantization levels (b) with NNA, and (c) with STE.}
  \label{LeNet5_acc_plots}
\end{figure*}

\subsection{Training feedforward BA-MLW NN: LeNet5 for MNIST}
\label{MNIST}
To demonstrate the \textcolor{\ReRevColor}{generalizability of our NNA algorithm to feedforward networks}
, we use the MNIST dataset and train BA-MLW networks using the LeNet5 architecture that comprises 2 CNN layers followed by 3 FC layers \cite{LeNet5-paper}. We compare the accuracy and noise resilience of the FP baseline with our NNA algorithm and STE, and the results are shown in Figure 5. Both STE and NNA are resilient to binarizing activations, and achieve peak accuracy comparable to the FP network and tolerate a wide range of $\sigma_\text{eval}$, with STE slightly outperforming NNA (Figure 5a). Both STE and NNA are also resilient to weight quantization (Figure 5b and 5c), with no loss of accuracy when quantizing weights down to 15 levels (a pair of 8-level cells), 
but slight accuracy degradation with 7-level weights. Compared with the GRU in Section~\ref{TWD}, LeNet5 needs more quantization levels due to its wider weight distribution ranges and smaller numbers of parameters in the CNN layers (especially the first CNN layer). 
Although not elaborated in this paper, layer-wise customized choices of quantization levels should further optimize performance.  

\textcolor{\ReRevColor}{Results of these two case studies show that the effectiveness of STE on feedforward networks does not easily translate to RNNs. }
While on the other hand, our NNA algorithm works well for both RNN and feedforward networks.


\section{Related work} 
Most prior quantization studies have been focused on feedforward networks, whereas quantizing RNNs turns out to be more challenging. Consistent with our results, quantization techniques that work well for feedforward NNs (e.g., STE) have been found to work poorly for RNNs \cite{QNN}. Existing RNN quantization studies find that to maintain accuracy, more bits are required for RNNs than for feedforward networks, especially for the activations. Hence, prior work either use FP activations \cite{RNN-QW-FPA}, or need multiple bits per activation \cite{RNN-QW-QA,QNN}, which would require costly DACs and ADCs for PIM implementations. In contrast, our work not only quantizes the weights but also binarizes activations of RNNs, enabling the optimal BA-MLW RNN structure for efficient PIM implementation.

Our NNA algorithm is largely inspired by the reparameterization trick \cite{VAE-Rezende} and the Gumbel-softmax trick \cite{Gumbel-Maddison,Gumbel-Eric}. Introduced in the context of variational inference, the reparameterization trick reformulates the sampling process of certain probability distributions (e.g., those having a ``location-scale'' form), which allows the expected gradient w.r.t.\ parameters of these distributions to propagate. Gumbel-softmax uses the Gumbel RVs to attain an equivalent sampling process from categorical distributions. Moreover, it uses a continuous relaxation trick to solve the gradient propagation problem of sampling from discrete distributions. \cite{1994-MLP-noise,1996-RNN-noise,1996-noise-generalization,1995-similarities} study the generalization effects of noise injection to NNs' inputs, weights, or activations. Additive Gaussian noise has also been used for learning binary encodings of documents with a multi-layer feedforward autoencoder \cite{Semantic-Hashing}. Our paper differentiates from these works in that we apply these techniques (noise injection and methods of propagating gradients through stochastic sampling nodes) to training BA-MLW RNNs in order to yield an optimal PIM circuit implementation that obviates DACs and ADCs. Moreover, we propose an effective noise annealing procedure in our NNA algorithm and use noise injection's regularization penalty effects to explain why our new algorithm enables high resilience to quantization and noise.


\textcolor{\RevisionColor}{A notable recent work that proposes an end-to-end analog NN implementation also strives to address the issues of AD/DA overhead and device/circuit non-idealities that have been plaguing NNs' PIM implementations \cite{EETimes,2020ANN}. Their solution exclusively applies to energy-based NN models that leverage the physical Kirchhoff's current law complied by a memristive crossbar network to find the corresponding NN model's mathematical minimal energy solution that is naturally represented by neurons' analog voltages, thereby avoiding ADC or DAC for hidden neurons during inference. To tackle device/circuit non-idealities, they adopt ``chip-in-the-loop'' training to tailor the NN model to each individual chip's variability after fabrication. In contrast, our approach targets the commonly used feedforward and recurrent NN models, rather than energy-base models. The BA-MLW NN models trained with our proposed NNA algorithm not only eliminate AD/DA overhead, but also are resilient to a wide range and variety of hardware noise profiles. As we have shown in the hardware validation section, the same pre-trained model can achieve high performance across different chips, which avoids the overhead of post-fabrication on-chip training or calibration.
}

\textcolor{\RevisionColor}{\section{Conclusions and Future Work}}
\textcolor{\RevisionColor}{There are two critical road blocks towards efficient PIM implementations of NN inference: the overwhelming power, area, and speed overhead from peripheral AD/DA circuitry, and inference accuracy degradation due to device and circuit non-idealities. We propose solving both problems by co-designing highly noise-resilient BA-MLW NN models (whose BAs obviate ADCs and DACs), trained using our novel NNA algorithm. The proposed noise injection and annealing based training procedure endows our NNs with not only high resilience to heavy quantizations, but also strong robustness against a variety of noise sources. Compared with a FP baseline and an alternative quantization algorithm (i.e., STE), our NNA algorithm achieves superior accuracy and noise resilience especially when applied to RNNs.} 

We demonstrate the architectural and circuit designs of a trigger word detecting PIM accelerator that implements a BA-MLW GRU trained with our NNA algorithm, and design detailed circuit noise models to evaluate its impact on inference performance. Assisted with a simple yet effective offset polarity random flipping circuit technique, our NNs maintain software-equivalent inference accuracy in the presence of the wide range and variety of noise encountered in real PIM circuits, revealing our NNs' surprisingly strong resilience to noise profiles even beyond the additive Gaussian RVs with which they are trained. Our proposed circuit and algorithm co-design strategies can help pave the path towards more efficient PIM implementations of NNs.

\endgroup


\newpage
\Urlmuskip=0mu plus 1mu\relax 
\bibliographystyle{ACM-Reference-Format}
\bibliography{mlsys2020_paper}


\begin{thebibliography}{40}


\ifx \showCODEN    \undefined \def \showCODEN     #1{\unskip}     \fi
\ifx \showDOI      \undefined \def \showDOI       #1{#1}\fi
\ifx \showISBNx    \undefined \def \showISBNx     #1{\unskip}     \fi
\ifx \showISBNxiii \undefined \def \showISBNxiii  #1{\unskip}     \fi
\ifx \showISSN     \undefined \def \showISSN      #1{\unskip}     \fi
\ifx \showLCCN     \undefined \def \showLCCN      #1{\unskip}     \fi
\ifx \shownote     \undefined \def \shownote      #1{#1}          \fi
\ifx \showarticletitle \undefined \def \showarticletitle #1{#1}   \fi
\ifx \showURL      \undefined \def \showURL       {\relax}        \fi
\providecommand\bibfield[2]{#2}
\providecommand\bibinfo[2]{#2}
\providecommand\natexlab[1]{#1}
\providecommand\showeprint[2][]{arXiv:#2}

\bibitem[\protect\citeauthoryear{An}{An}{1996}]%
        {1996-noise-generalization}
\bibfield{author}{\bibinfo{person}{G. An}.} \bibinfo{year}{1996}\natexlab{}.
\newblock \showarticletitle{The effects of adding noise during backpropagation
  training on a generalization performance}.
\newblock \bibinfo{journal}{\emph{Neural computation}} \bibinfo{volume}{8},
  \bibinfo{number}{3} (\bibinfo{year}{1996}), \bibinfo{pages}{643--674}.
\newblock


\bibitem[\protect\citeauthoryear{{Bankman}, {Yang}, {Moons}, {Verhelst}, and
  {Murmann}}{{Bankman} et~al\mbox{.}}{2019}]%
        {2018-ISSCC-SAR-VOS}
\bibfield{author}{\bibinfo{person}{D. {Bankman}}, \bibinfo{person}{L. {Yang}},
  \bibinfo{person}{B. {Moons}}, \bibinfo{person}{M. {Verhelst}}, {and}
  \bibinfo{person}{B. {Murmann}}.} \bibinfo{year}{2019}\natexlab{}.
\newblock \showarticletitle{An Always-On 3.8 $\mu$ {J}/86\% {CIFAR}-10
  Mixed-Signal Binary {CNN} Processor With All Memory on Chip in 28-nm {CMOS}}.
\newblock \bibinfo{journal}{\emph{IEEE Journal of Solid-State Circuits}}
  \bibinfo{volume}{54}, \bibinfo{number}{1} (\bibinfo{year}{2019}),
  \bibinfo{pages}{158--172}.
\newblock


\bibitem[\protect\citeauthoryear{Bedeschi et~al\mbox{.}}{Bedeschi
  et~al\mbox{.}}{2008}]%
        {PCM}
\bibfield{author}{\bibinfo{person}{F. Bedeschi} {et~al\mbox{.}}}
  \bibinfo{year}{2008}\natexlab{}.
\newblock \showarticletitle{A bipolar-selected phase change memory featuring
  multi-level cell storage}.
\newblock \bibinfo{journal}{\emph{IEEE JSSC}} \bibinfo{volume}{44},
  \bibinfo{number}{1} (\bibinfo{year}{2008}), \bibinfo{pages}{217--227}.
\newblock


\bibitem[\protect\citeauthoryear{Bengio et~al\mbox{.}}{Bengio
  et~al\mbox{.}}{2013}]%
        {Bengio2013}
\bibfield{author}{\bibinfo{person}{Y. Bengio} {et~al\mbox{.}}}
  \bibinfo{year}{2013}\natexlab{}.
\newblock \showarticletitle{Estimating or propagating gradients through
  stochastic neurons for conditional computation}.
\newblock \bibinfo{journal}{\emph{arXiv:1308.3432}} (\bibinfo{year}{2013}).
\newblock


\bibitem[\protect\citeauthoryear{Bennett, Xiao, Dellana, Feinberg, Agarwal,
  Marinella, Agrawal, Prabhakar, Ramkumar, Hinh, et~al\mbox{.}}{Bennett
  et~al\mbox{.}}{2020}]%
        {2020-IRPS}
\bibfield{author}{\bibinfo{person}{Christopher~H Bennett},
  \bibinfo{person}{T~Patrick Xiao}, \bibinfo{person}{Ryan Dellana},
  \bibinfo{person}{Ben Feinberg}, \bibinfo{person}{Sapan Agarwal},
  \bibinfo{person}{Matthew~J Marinella}, \bibinfo{person}{Vineet Agrawal},
  \bibinfo{person}{Venkatraman Prabhakar}, \bibinfo{person}{Krishnaswamy
  Ramkumar}, \bibinfo{person}{Long Hinh}, {et~al\mbox{.}}}
  \bibinfo{year}{2020}\natexlab{}.
\newblock \showarticletitle{Device-aware inference operations in {SONOS}
  nonvolatile memory arrays}. In \bibinfo{booktitle}{\emph{2020 IEEE
  International Reliability Physics Symposium (IRPS)}}. IEEE,
  \bibinfo{pages}{1--6}.
\newblock


\bibitem[\protect\citeauthoryear{Cho et~al\mbox{.}}{Cho et~al\mbox{.}}{2014}]%
        {GRU-paper}
\bibfield{author}{\bibinfo{person}{K. Cho} {et~al\mbox{.}}}
  \bibinfo{year}{2014}\natexlab{}.
\newblock \showarticletitle{Learning phrase representations using {RNN}
  encoder-decoder for statistical machine translation}.
\newblock \bibinfo{journal}{\emph{arXiv:1406.1078}} (\bibinfo{year}{2014}).
\newblock


\bibitem[\protect\citeauthoryear{Chung et~al\mbox{.}}{Chung
  et~al\mbox{.}}{2016}]%
        {STE-slope-anneal}
\bibfield{author}{\bibinfo{person}{J. Chung} {et~al\mbox{.}}}
  \bibinfo{year}{2016}\natexlab{}.
\newblock \showarticletitle{Hierarchical multiscale recurrent neural networks}.
\newblock \bibinfo{journal}{\emph{arXiv:1609.01704}} (\bibinfo{year}{2016}).
\newblock


\bibitem[\protect\citeauthoryear{Fick et~al\mbox{.}}{Fick
  et~al\mbox{.}}{2017}]%
        {SONOS-PIM}
\bibfield{author}{\bibinfo{person}{L. Fick} {et~al\mbox{.}}}
  \bibinfo{year}{2017}\natexlab{}.
\newblock \showarticletitle{Analog in-memory subthreshold deep neural network
  accelerator}. In \bibinfo{booktitle}{\emph{IEEE CICC}}.
  \bibinfo{pages}{1--4}.
\newblock


\bibitem[\protect\citeauthoryear{He et~al\mbox{.}}{He et~al\mbox{.}}{2016}]%
        {RNN-QW-QA}
\bibfield{author}{\bibinfo{person}{Q. He} {et~al\mbox{.}}}
  \bibinfo{year}{2016}\natexlab{}.
\newblock \showarticletitle{Effective quantization methods for recurrent neural
  networks}.
\newblock \bibinfo{journal}{\emph{arXiv:1611.10176}} (\bibinfo{year}{2016}).
\newblock


\bibitem[\protect\citeauthoryear{He, Lin, Ewetz, Yuan, and Fan}{He
  et~al\mbox{.}}{2019}]%
        {2019-DAC}
\bibfield{author}{\bibinfo{person}{Zhezhi He}, \bibinfo{person}{Jie Lin},
  \bibinfo{person}{Rickard Ewetz}, \bibinfo{person}{Jiann-Shiun Yuan}, {and}
  \bibinfo{person}{Deliang Fan}.} \bibinfo{year}{2019}\natexlab{}.
\newblock \showarticletitle{Noise injection adaption: End-to-end {R}e{RAM}
  crossbar non-ideal effect adaption for neural network mapping}. In
  \bibinfo{booktitle}{\emph{Proceedings of the 56th Annual Design Automation
  Conference 2019}}. \bibinfo{pages}{1--6}.
\newblock


\bibitem[\protect\citeauthoryear{Hubara et~al\mbox{.}}{Hubara
  et~al\mbox{.}}{2017}]%
        {QNN}
\bibfield{author}{\bibinfo{person}{I. Hubara} {et~al\mbox{.}}}
  \bibinfo{year}{2017}\natexlab{}.
\newblock \showarticletitle{Quantized neural networks: Training neural networks
  with low precision weights and activations}.
\newblock \bibinfo{journal}{\emph{The Journal of Machine Learning Research}}
  \bibinfo{volume}{18}, \bibinfo{number}{1} (\bibinfo{year}{2017}),
  \bibinfo{pages}{6869--6898}.
\newblock


\bibitem[\protect\citeauthoryear{Jang et~al\mbox{.}}{Jang
  et~al\mbox{.}}{2016}]%
        {Gumbel-Eric}
\bibfield{author}{\bibinfo{person}{E. Jang} {et~al\mbox{.}}}
  \bibinfo{year}{2016}\natexlab{}.
\newblock \showarticletitle{Categorical reparameterization with
  {G}umbel-softmax}.
\newblock \bibinfo{journal}{\emph{arXiv:1611.01144}} (\bibinfo{year}{2016}).
\newblock


\bibitem[\protect\citeauthoryear{Jim et~al\mbox{.}}{Jim et~al\mbox{.}}{1996}]%
        {1996-RNN-noise}
\bibfield{author}{\bibinfo{person}{K.-C. Jim} {et~al\mbox{.}}}
  \bibinfo{year}{1996}\natexlab{}.
\newblock \showarticletitle{An analysis of noise in recurrent neural networks:
  convergence and generalization}.
\newblock \bibinfo{journal}{\emph{IEEE Transactions on neural networks}}
  \bibinfo{volume}{7}, \bibinfo{number}{6} (\bibinfo{year}{1996}),
  \bibinfo{pages}{1424--1438}.
\newblock


\bibitem[\protect\citeauthoryear{Joshi, Le~Gallo, Haefeli, Boybat, Nandakumar,
  Piveteau, Dazzi, Rajendran, Sebastian, and Eleftheriou}{Joshi
  et~al\mbox{.}}{2020}]%
        {2020-Nature}
\bibfield{author}{\bibinfo{person}{Vinay Joshi}, \bibinfo{person}{Manuel
  Le~Gallo}, \bibinfo{person}{Simon Haefeli}, \bibinfo{person}{Irem Boybat},
  \bibinfo{person}{Sasidharan~Rajalekshmi Nandakumar},
  \bibinfo{person}{Christophe Piveteau}, \bibinfo{person}{Martino Dazzi},
  \bibinfo{person}{Bipin Rajendran}, \bibinfo{person}{Abu Sebastian}, {and}
  \bibinfo{person}{Evangelos Eleftheriou}.} \bibinfo{year}{2020}\natexlab{}.
\newblock \showarticletitle{Accurate deep neural network inference using
  computational phase-change memory}.
\newblock \bibinfo{journal}{\emph{Nature Communications}} \bibinfo{volume}{11},
  \bibinfo{number}{1} (\bibinfo{year}{2020}), \bibinfo{pages}{1--13}.
\newblock


\bibitem[\protect\citeauthoryear{Kendall, Pantone, Manickavasagam, Bengio, and
  Scellier}{Kendall et~al\mbox{.}}{2020}]%
        {2020ANN}
\bibfield{author}{\bibinfo{person}{Jack Kendall}, \bibinfo{person}{Ross
  Pantone}, \bibinfo{person}{Kalpana Manickavasagam}, \bibinfo{person}{Yoshua
  Bengio}, {and} \bibinfo{person}{Benjamin Scellier}.}
  \bibinfo{year}{2020}\natexlab{}.
\newblock \bibinfo{title}{Training End-to-End Analog Neural Networks with
  Equilibrium Propagation}.
\newblock
\newblock
\showeprint[arxiv]{cs.NE/2006.01981}


\bibitem[\protect\citeauthoryear{Khwa et~al\mbox{.}}{Khwa
  et~al\mbox{.}}{2018}]%
        {PIM-BNN-SenseAmp}
\bibfield{author}{\bibinfo{person}{W.-S. Khwa} {et~al\mbox{.}}}
  \bibinfo{year}{2018}\natexlab{}.
\newblock \showarticletitle{A 65nm 4{K}b algorithm-dependent
  computing-in-memory {SRAM} unit-macro with 2.3 ns and 55.8 {TOPS}/{W} fully
  parallel product-sum operation for binary {DNN} edge processors}. In
  \bibinfo{booktitle}{\emph{IEEE ISSCC}}. \bibinfo{pages}{496--498}.
\newblock


\bibitem[\protect\citeauthoryear{Klachko, Mahmoodi, and Strukov}{Klachko
  et~al\mbox{.}}{2019}]%
        {2019-UCSB-IJCNN-ShotNoise}
\bibfield{author}{\bibinfo{person}{Michael Klachko},
  \bibinfo{person}{Mohammad~Reza Mahmoodi}, {and} \bibinfo{person}{Dmitri
  Strukov}.} \bibinfo{year}{2019}\natexlab{}.
\newblock \showarticletitle{Improving noise tolerance of mixed-signal neural
  networks}. In \bibinfo{booktitle}{\emph{2019 International Joint Conference
  on Neural Networks (IJCNN)}}. IEEE, \bibinfo{pages}{1--8}.
\newblock


\bibitem[\protect\citeauthoryear{LeCun et~al\mbox{.}}{LeCun
  et~al\mbox{.}}{1998}]%
        {LeNet5-paper}
\bibfield{author}{\bibinfo{person}{Y. LeCun} {et~al\mbox{.}}}
  \bibinfo{year}{1998}\natexlab{}.
\newblock \showarticletitle{Gradient-based learning applied to document
  recognition}.
\newblock \bibinfo{journal}{\emph{Proc. IEEE}} \bibinfo{volume}{86},
  \bibinfo{number}{11} (\bibinfo{year}{1998}), \bibinfo{pages}{2278--2324}.
\newblock


\bibitem[\protect\citeauthoryear{Li et~al\mbox{.}}{Li et~al\mbox{.}}{2016}]%
        {Ternary-net}
\bibfield{author}{\bibinfo{person}{F. Li} {et~al\mbox{.}}}
  \bibinfo{year}{2016}\natexlab{}.
\newblock \showarticletitle{Ternary weight networks}.
\newblock \bibinfo{journal}{\emph{arXiv:1605.04711}} (\bibinfo{year}{2016}).
\newblock


\bibitem[\protect\citeauthoryear{Ma, Donato, Lee, Brooks, and Wei}{Ma
  et~al\mbox{.}}{2019}]%
        {CMOS-MLC}
\bibfield{author}{\bibinfo{person}{Siming Ma}, \bibinfo{person}{Marco Donato},
  \bibinfo{person}{Sae~Kyu Lee}, \bibinfo{person}{David Brooks}, {and}
  \bibinfo{person}{Gu-Yeon Wei}.} \bibinfo{year}{2019}\natexlab{}.
\newblock \showarticletitle{Fully-{CMOS} Multi-Level Embedded Non-Volatile
  Memory Devices With Reliable Long-Term Retention for Efficient Storage of
  Neural Network Weights}.
\newblock \bibinfo{journal}{\emph{IEEE Electron Device Letters}}
  \bibinfo{volume}{40}, \bibinfo{number}{9} (\bibinfo{year}{2019}),
  \bibinfo{pages}{1403--1406}.
\newblock


\bibitem[\protect\citeauthoryear{Maddison et~al\mbox{.}}{Maddison
  et~al\mbox{.}}{2016}]%
        {Gumbel-Maddison}
\bibfield{author}{\bibinfo{person}{C.~J. Maddison} {et~al\mbox{.}}}
  \bibinfo{year}{2016}\natexlab{}.
\newblock \showarticletitle{The concrete distribution: A continuous relaxation
  of discrete random variables}.
\newblock \bibinfo{journal}{\emph{arXiv:1611.00712}} (\bibinfo{year}{2016}).
\newblock


\bibitem[\protect\citeauthoryear{Merolla, Appuswamy, Arthur, Esser, and
  Modha}{Merolla et~al\mbox{.}}{2016}]%
        {2016-TolerantVariety}
\bibfield{author}{\bibinfo{person}{Paul Merolla}, \bibinfo{person}{Rathinakumar
  Appuswamy}, \bibinfo{person}{John Arthur}, \bibinfo{person}{Steve~K Esser},
  {and} \bibinfo{person}{Dharmendra Modha}.} \bibinfo{year}{2016}\natexlab{}.
\newblock \showarticletitle{Deep neural networks are robust to weight
  binarization and other non-linear distortions}.
\newblock \bibinfo{journal}{\emph{arXiv preprint arXiv:1606.01981}}
  (\bibinfo{year}{2016}).
\newblock


\bibitem[\protect\citeauthoryear{Miyahara, Asada, Paik, and Matsuzawa}{Miyahara
  et~al\mbox{.}}{2008}]%
        {SenseAmp}
\bibfield{author}{\bibinfo{person}{Masaya Miyahara}, \bibinfo{person}{Yusuke
  Asada}, \bibinfo{person}{Daehwa Paik}, {and} \bibinfo{person}{Akira
  Matsuzawa}.} \bibinfo{year}{2008}\natexlab{}.
\newblock \showarticletitle{A low-noise self-calibrating dynamic comparator for
  high-speed {ADC}s}. In \bibinfo{booktitle}{\emph{2008 IEEE Asian Solid-State
  Circuits Conference}}. IEEE, \bibinfo{pages}{269--272}.
\newblock


\bibitem[\protect\citeauthoryear{Moon, Shin, and Jeon}{Moon
  et~al\mbox{.}}{2019}]%
        {2019-TVLSI-SAR-VOS}
\bibfield{author}{\bibinfo{person}{Suhong Moon}, \bibinfo{person}{Kwanghyun
  Shin}, {and} \bibinfo{person}{Dongsuk Jeon}.}
  \bibinfo{year}{2019}\natexlab{}.
\newblock \showarticletitle{Enhancing reliability of analog neural network
  processors}.
\newblock \bibinfo{journal}{\emph{IEEE Transactions on Very Large Scale
  Integration (VLSI) Systems}} \bibinfo{volume}{27}, \bibinfo{number}{6}
  (\bibinfo{year}{2019}), \bibinfo{pages}{1455--1459}.
\newblock


\bibitem[\protect\citeauthoryear{Murray et~al\mbox{.}}{Murray
  et~al\mbox{.}}{1994}]%
        {1994-MLP-noise}
\bibfield{author}{\bibinfo{person}{A. Murray} {et~al\mbox{.}}}
  \bibinfo{year}{1994}\natexlab{}.
\newblock \showarticletitle{Enhanced {MLP} performance and fault tolerance
  resulting from synaptic weight noise during training}.
\newblock \bibinfo{journal}{\emph{IEEE Transactions on neural networks}}
  \bibinfo{volume}{5}, \bibinfo{number}{5} (\bibinfo{year}{1994}),
  \bibinfo{pages}{792--802}.
\newblock


\bibitem[\protect\citeauthoryear{Ng et~al\mbox{.}}{Ng et~al\mbox{.}}{2018}]%
        {Andrew-Ng-GRU}
\bibfield{author}{\bibinfo{person}{A. Ng} {et~al\mbox{.}}}
  \bibinfo{year}{2018}\natexlab{}.
\newblock \bibinfo{title}{Recurrent neural network: gated recurrent unit
  ({GRU})}.
\newblock
\newblock
\urldef\tempurl%
\url{https://www.youtube.com/watch?v=xSCy3q2ts44}
\showURL{%
\tempurl}


\bibitem[\protect\citeauthoryear{Ott et~al\mbox{.}}{Ott et~al\mbox{.}}{2016}]%
        {RNN-QW-FPA}
\bibfield{author}{\bibinfo{person}{J. Ott} {et~al\mbox{.}}}
  \bibinfo{year}{2016}\natexlab{}.
\newblock \showarticletitle{Recurrent neural networks with limited numerical
  precision}.
\newblock \bibinfo{journal}{\emph{arXiv:1608.06902}} (\bibinfo{year}{2016}).
\newblock


\bibitem[\protect\citeauthoryear{Rastegari et~al\mbox{.}}{Rastegari
  et~al\mbox{.}}{2016}]%
        {XNOR-net}
\bibfield{author}{\bibinfo{person}{M. Rastegari} {et~al\mbox{.}}}
  \bibinfo{year}{2016}\natexlab{}.
\newblock \showarticletitle{Xnor-net: {I}magenet classification using binary
  convolutional neural networks}. In \bibinfo{booktitle}{\emph{European
  Conference on Computer Vision}}. Springer, \bibinfo{pages}{525--542}.
\newblock


\bibitem[\protect\citeauthoryear{Reed et~al\mbox{.}}{Reed
  et~al\mbox{.}}{1995}]%
        {1995-similarities}
\bibfield{author}{\bibinfo{person}{R. Reed} {et~al\mbox{.}}}
  \bibinfo{year}{1995}\natexlab{}.
\newblock \showarticletitle{Similarities of error regularization, sigmoid gain
  scaling, target smoothing, and training with jitter}.
\newblock \bibinfo{journal}{\emph{IEEE Transactions on Neural Networks}}
  \bibinfo{volume}{6}, \bibinfo{number}{3} (\bibinfo{year}{1995}),
  \bibinfo{pages}{529--538}.
\newblock


\bibitem[\protect\citeauthoryear{Rezende et~al\mbox{.}}{Rezende
  et~al\mbox{.}}{2014}]%
        {VAE-Rezende}
\bibfield{author}{\bibinfo{person}{D. Rezende} {et~al\mbox{.}}}
  \bibinfo{year}{2014}\natexlab{}.
\newblock \showarticletitle{Stochastic backpropagation and approximate
  inference in deep generative models}.
\newblock \bibinfo{journal}{\emph{arXiv:1401.4082}} (\bibinfo{year}{2014}).
\newblock


\bibitem[\protect\citeauthoryear{Salakhutdinov et~al\mbox{.}}{Salakhutdinov
  et~al\mbox{.}}{2009}]%
        {Semantic-Hashing}
\bibfield{author}{\bibinfo{person}{R. Salakhutdinov} {et~al\mbox{.}}}
  \bibinfo{year}{2009}\natexlab{}.
\newblock \showarticletitle{Semantic hashing}.
\newblock \bibinfo{journal}{\emph{International Journal of Approximate
  Reasoning}} \bibinfo{volume}{50}, \bibinfo{number}{7} (\bibinfo{year}{2009}),
  \bibinfo{pages}{969--978}.
\newblock


\bibitem[\protect\citeauthoryear{Shafiee et~al\mbox{.}}{Shafiee
  et~al\mbox{.}}{2016}]%
        {isaac}
\bibfield{author}{\bibinfo{person}{A. Shafiee} {et~al\mbox{.}}}
  \bibinfo{year}{2016}\natexlab{}.
\newblock \showarticletitle{{ISAAC}: A convolutional neural network accelerator
  with in-situ analog arithmetic in crossbars}.
\newblock \bibinfo{journal}{\emph{ACM SIGARCH Computer Architecture News}}
  \bibinfo{volume}{44}, \bibinfo{number}{3} (\bibinfo{year}{2016}),
  \bibinfo{pages}{14--26}.
\newblock


\bibitem[\protect\citeauthoryear{Sheu et~al\mbox{.}}{Sheu
  et~al\mbox{.}}{2011}]%
        {ReRAM}
\bibfield{author}{\bibinfo{person}{S.-S. Sheu} {et~al\mbox{.}}}
  \bibinfo{year}{2011}\natexlab{}.
\newblock \showarticletitle{A 4{M}b embedded {SLC} resistive-{RAM} macro with
  7.2 ns read-write random-access time and 160ns {MLC}-access capability}. In
  \bibinfo{booktitle}{\emph{IEEE ISSCC}}. \bibinfo{pages}{200--202}.
\newblock


\bibitem[\protect\citeauthoryear{Song et~al\mbox{.}}{Song
  et~al\mbox{.}}{2017}]%
        {pipelayer-serial-AD-DA}
\bibfield{author}{\bibinfo{person}{L. Song} {et~al\mbox{.}}}
  \bibinfo{year}{2017}\natexlab{}.
\newblock \showarticletitle{Pipelayer: A pipelined {R}e{RAM}-based accelerator
  for deep learning}. In \bibinfo{booktitle}{\emph{IEEE HPCA}}.
  \bibinfo{pages}{541--552}.
\newblock


\bibitem[\protect\citeauthoryear{Ward-Foxton}{Ward-Foxton}{2020}]%
        {EETimes}
\bibfield{author}{\bibinfo{person}{Sally Ward-Foxton}.}
  \bibinfo{year}{2020}\natexlab{}.
\newblock \showarticletitle{Research Proves End-to-End Analog Chips for {AI}
  Computation Possible}.
\newblock \bibinfo{journal}{\emph{EETimes}} (\bibinfo{year}{2020}).
\newblock
\urldef\tempurl%
\url{https://www.eetimes.com/research-breakthrough-promises-end-to-end-analog-chips-for-ai-computation/}
\showURL{%
\tempurl}


\bibitem[\protect\citeauthoryear{Warden}{Warden}{2018}]%
        {Speech-Commands-Warden}
\bibfield{author}{\bibinfo{person}{P. Warden}.}
  \bibinfo{year}{2018}\natexlab{}.
\newblock \showarticletitle{Speech commands: A dataset for limited-vocabulary
  speech recognition}.
\newblock \bibinfo{journal}{\emph{arXiv:1804.03209}} (\bibinfo{year}{2018}).
\newblock


\bibitem[\protect\citeauthoryear{Wu et~al\mbox{.}}{Wu et~al\mbox{.}}{2014}]%
        {TSMC-16nm}
\bibfield{author}{\bibinfo{person}{S.-Y. Wu} {et~al\mbox{.}}}
  \bibinfo{year}{2014}\natexlab{}.
\newblock \showarticletitle{An enhanced 16nm {CMOS} technology featuring 2 nd
  generation {F}in{FET} transistors and advanced {C}u/low-k interconnect for
  low power and high performance applications}. In
  \bibinfo{booktitle}{\emph{IEEE IEDM}}. IEEE, \bibinfo{pages}{3--1}.
\newblock


\bibitem[\protect\citeauthoryear{Xiao, Bennett, Feinberg, Agarwal, and
  Marinella}{Xiao et~al\mbox{.}}{2020}]%
        {2020Review}
\bibfield{author}{\bibinfo{person}{T~Patrick Xiao},
  \bibinfo{person}{Christopher~H Bennett}, \bibinfo{person}{Ben Feinberg},
  \bibinfo{person}{Sapan Agarwal}, {and} \bibinfo{person}{Matthew~J
  Marinella}.} \bibinfo{year}{2020}\natexlab{}.
\newblock \showarticletitle{Analog architectures for neural network
  acceleration based on non-volatile memory}.
\newblock \bibinfo{journal}{\emph{Applied Physics Reviews}}
  \bibinfo{volume}{7}, \bibinfo{number}{3} (\bibinfo{year}{2020}),
  \bibinfo{pages}{031301}.
\newblock


\bibitem[\protect\citeauthoryear{Yang and Sze}{Yang and Sze}{2019}]%
        {2019IEDM}
\bibfield{author}{\bibinfo{person}{Tien-Ju Yang} {and}
  \bibinfo{person}{Vivienne Sze}.} \bibinfo{year}{2019}\natexlab{}.
\newblock \showarticletitle{Design Considerations for Efficient Deep Neural
  Networks on Processing-in-Memory Accelerators}. In
  \bibinfo{booktitle}{\emph{2019 IEEE International Electron Devices Meeting
  (IEDM)}}. IEEE, \bibinfo{pages}{22--1}.
\newblock


\bibitem[\protect\citeauthoryear{Zhang et~al\mbox{.}}{Zhang
  et~al\mbox{.}}{2016}]%
        {Naveen-BL-DAC}
\bibfield{author}{\bibinfo{person}{J. Zhang} {et~al\mbox{.}}}
  \bibinfo{year}{2016}\natexlab{}.
\newblock \showarticletitle{A machine-learning classifier implemented in a
  standard 6{T} {SRAM} array}. In \bibinfo{booktitle}{\emph{IEEE
  VLSI-Circuits}}. \bibinfo{pages}{1--2}.
\newblock


\end{thebibliography}

\end{document}